\documentclass[runningheads]{llncs}
\usepackage[T1]{fontenc}

\usepackage[utf8]{inputenc}
\usepackage{multirow}
\usepackage{booktabs}
\usepackage{array}
\usepackage{graphicx}
\usepackage{caption}
\usepackage{makecell} 
\usepackage{amsmath}
\begin{document}
\title{FedGIN: Federated Learning with Dynamic Global Intensity Non-linear Augmentation for Organ Segmentation using Multi-modal Images}
%
\author{Sachin Dudda Nagaraju\inst{1}\and Ashkan Moradi \inst{1}\and
Bendik Skarre Abrahamsen\inst{1} \and
Mattijs Elschot\inst{1,2}}
\authorrunning{SDN. Author et al.}
%
\institute{Department of Circulation and Medical Imaging, Norwegian University of Science and Technology, Trondheim, Norway \and
Central Staff, St. Olavs Hospital, Trondheim University Hospital, Trondheim, Norway\\
\email{\{sachin.d.nagaraju,ashkan.moradi,bendik.s.abrahamsen,mattijs.elschot\}@ntnu.no}}


\maketitle              
\begin{abstract}
Medical image segmentation plays a crucial role in AI-assisted diagnostics, surgical planning, and treatment monitoring. Accurate and robust segmentation models are essential for enabling reliable, data-driven clinical decision making across diverse imaging modalities. Given the inherent variability in image characteristics across modalities, developing a unified model capable of generalizing effectively to multiple modalities would be highly beneficial. This model could streamline clinical workflows and reduce the need for modality-specific training. However, real-world deployment faces major challenges, including data scarcity, domain shift between modalities (e.g., CT vs. MRI), and privacy restrictions that prevent data sharing. To address these issues, we propose FedGIN, a Federated Learning (FL) framework that enables multimodal organ segmentation without sharing raw patient data. Our method integrates a lightweight Global Intensity Non-linear (GIN) augmentation module that harmonizes modality-specific intensity distributions during local training. We evaluated FedGIN using two types of datasets: a \textit{limited dataset} and a \textit{complete dataset}. In the limited dataset scenario, the model was initially trained using only MRI data, and CT data was added to assess its performance improvements. In the complete dataset scenario, both MRI and CT data were fully utilized for training on all clients. In the limited-data scenario, FedGIN achieved a 12–18\% improvement in 3D Dice scores on MRI test cases compared to FL without GIN and consistently outperformed local baselines. In the complete dataset scenario, FedGIN demonstrated near-centralized performance, with a 30\% Dice score improvement over the MRI-only baseline and a 10\% improvement over the CT-only baseline, highlighting its strong cross-modality generalization under privacy constraints. Code available here https://github.com/sachugowda/FedGIN/.


\end{abstract}
\section{Introduction}

Artificial intelligence (AI) has revolutionized medical imaging by automating complex diagnostic analysis, enhancing accuracy, and improving clinical workflows. Deep learning methods, especially convolutional neural networks, have achieved expert-level performance in segmentation tasks across organs and modalities \cite{guan2024federated}. However, developing generalizable and robust models demands access to large datasets that are diverse in imaging modalities, anatomical regions, and patient populations. This requirement often conflicts with stringent privacy regulations and institutional governance policies, which restrict direct data sharing between sites \cite{pati2024privacy}. Multimodal medical imaging datasets encompassing modalities such as computed tomography (CT) and magnetic resonance imaging (MRI) are crucial for developing generalizable AI models, as each modality captures distinct anatomical features. Rather than combining modalities for individual clinical cases, a more scalable and practical objective is to train unified segmentation models capable of operating across different modalities \cite{lassau2020three}. These models benefit from the diversity of multimodal data while avoiding dependence on modality-specific architectures. However, developing these models faces a fundamental barrier: multimodal datasets are isolated across institutions, limiting the ability to leverage their collective value for robust model training \cite{pati2024privacy}.

Federated learning (FL) \cite{mcmahan2017communication} offers a solution for privacy and data accessibility in medical imaging by enabling collaborative model training across institutions without sharing raw data. Each client trains a local model on private data and shares only model parameters with a central server to build a global model \cite{yuan2024communication}. FL has been successfully applied in tasks like brain MRI analysis and MRI-to-CT synthesis for radiotherapy planning \cite{raggio2025fedsynthct}. However, challenges remain, including unpaired data (e.g., CT and MRI images from different patients), modality differences, and non-IID data \cite{zhao2018federated} across institutions, which complicate the deployment of multimodal models. Recent centralized multimodal methods \cite{d2024totalsegmentator,ciausu2024towards} improve segmentation through feature alignment. However, their reliance on complex architectures and extensive tuning limits scalability in resource-constrained clinical settings. Additionally, their performance in segmenting low-contrast, anatomically diverse organs is underexplored. Approaches like CAR-MFL \cite{wang2025multi} and other federated frameworks \cite{wang2025clusmfl,myrzashova2025bcftl} address these challenges, but most require paired multimodal images, which is often impractical due to clinical protocol variations. Developing robust, generalizable solutions for unpaired multimodal data, especially with CT and MRI, remains an open challenge due to domain shifts, scanner variations, and anatomical differences, leading to performance degradation and limiting cross-site, cross-modality model reliability. In these scenarios, leveraging complementary CT data through FL could enhance MRI-based segmentation, particularly for rare cancers and anatomically complex regions \cite{chen2025generalizable}. To mitigate cross-modality distribution differences, domain generalization strategies have recently gained traction. Global Intensity Non-linear (GIN) augmentation \cite{ouyang2022causality} has shown promise in bridging CT-MRI intensity differences under centralized settings, demonstrating the ability to train on one modality and generalize to another, its application has primarily focused on single-modality training and cross-modality testing. The potential of using GIN across multiple modalities during training to harmonize multimodal data in federated environments is underexplored. Addressing this could improve FL’s generalization across modalities and institutions, enabling scalable, privacy-preserving medical AI systems.

To address these critical gaps, we propose a novel FL framework that integrates GIN augmentation for domain generalization in organ segmentation tasks using unpaired multimodal image data.  Our approach specifically tackles the challenges of intensity distribution mismatches between CT and MRI data through causality-inspired augmentation techniques that expose models to synthesized domain-shifted training examples. The framework enables secure cross-modality learning under data scarcity by allowing institutions to contribute either unimodal or multimodal data while maintaining robust segmentation performance. We hypothesize that by dynamically harmonizing modality differences during training, the integration of GIN augmentation may enhance the model’s ability to learn invariant representations that generalize more effectively across unseen modalities and institutions.

\section{Methodology}

\begin{figure}[t]
\centering
\includegraphics[width=1.0\textwidth]{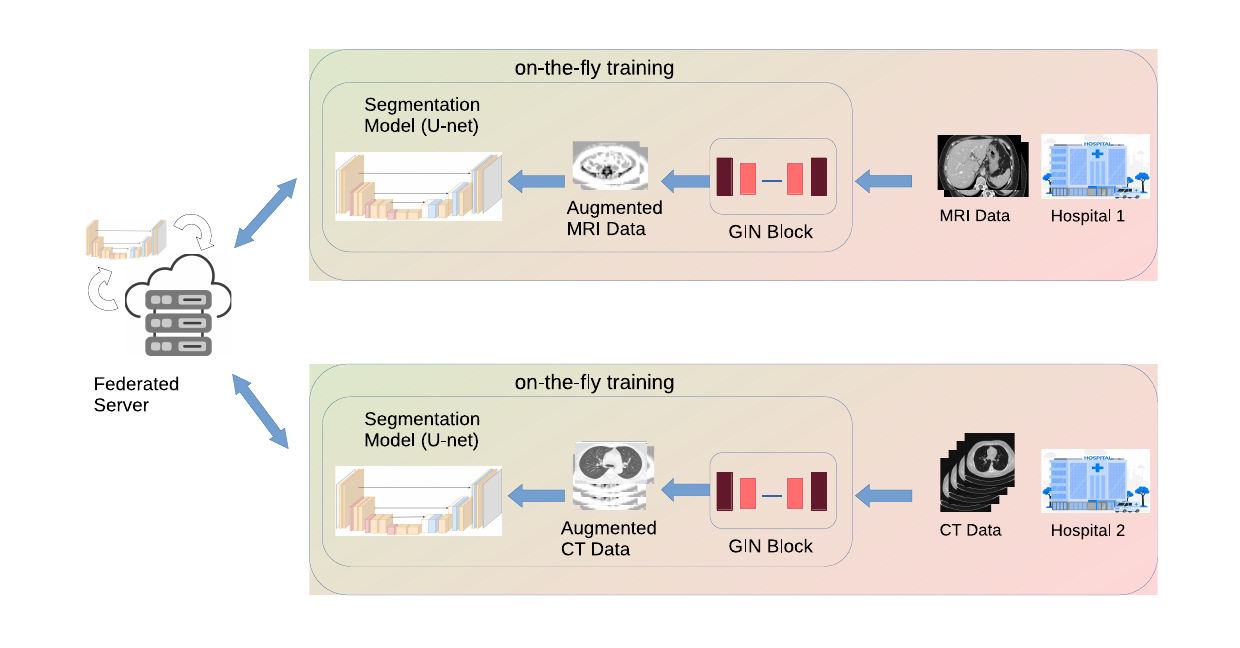}
\caption{Proposed FedGIN approach for FL with GIN augmentation for organ segmentation using multi-modal images} \label{fig1}
\vspace{-20pt} %
\end{figure}

We present FedGIN, a FL framework designed to build a generalizable organ segmentation model across institutions contributing different imaging modalities. Each client, which may possess unpaired data from MRI, CT, or both modalities, begins local training on its own data using the globally shared model received from the server. To facilitate cross-modality learning under these unpaired, modality-diverse conditions, we integrate a GIN augmentation module to effectively address modality-specific domain shifts; GIN applies randomized, anatomy-preserving intensity transformations during training. This encourages the model to learn modality-invariant features. After local training, model updates are sent to a central server for aggregation. This process is repeated over multiple communication rounds to iteratively refine a global model that generalizes across both modalities and different clients. 

\subsection{GIN Augmentation and Federated Workflow}

To address modality-induced domain shifts, we use a GIN \cite{ouyang2022causality} module that operates on-the-fly during local training. GIN uses shallow convolutional networks with weights sampled from a Gaussian distribution $\mathcal{N}(0, I)$ at each iteration. Leaky ReLU activations are applied between layers to introduce non-linearity, and no downsampling is used to preserve spatial resolution. The augmented image is generated by blending the shallow convolutional network's output $g^{\text{Net}}_\theta(x)$ with the original input $x$, using a random coefficient $\alpha \sim \mathcal{U}(0,1)$, followed by Frobenius norm normalization:
\[
g_\theta(x) = \alpha \cdot g^{\text{Net}}_\theta(x) + (1 - \alpha) \cdot x,\quad
g_\theta(x) \leftarrow \frac{\|x\|_F}{\|g_\theta(x)\|_F} \cdot g_\theta(x).
\]
This augmentation exposes the model to diverse intensity and texture variations while preserving structural content, thereby improving generalization.

The overall FL process begins with the server broadcasting the global model to all clients. Each client applies GIN-based augmentation on its local CT or MRI data and trains a U-Net model for several local epochs. After training, model weights are sent back to the server, which performs aggregation via FedAvg \cite{mcmahan2017communication}. The updated global model is redistributed, and this cycle continues for multiple communication rounds. The full pipeline is illustrated in Fig.~\ref{fig1}.

\section{Experiments}
To evaluate our method, we conducted experiments with a two-client FL setup: one client trained on CT data, the other on MRI. This allowed us to assess the model's generalization across modalities, with a centralized combination of MRI and CT data, and local models trained on individual modalities.

\subsection{Dataset details}

We conducted experiments using two public 3D medical imaging datasets: TotalSegmentator \cite{wasserthal2023totalsegmentator} and AMOS2020 \cite{ji2022amos}. The TotalSegmentator dataset, containing CT and MRI images, was used for training and validation, while the AMOS2020 dataset was reserved for testing. We aimed to evaluate generalization across modalities for five abdominal organs: liver, kidneys, spleen, pancreas, and gall bladder. Each 3D dataset was sliced along the axial plane to generate 2D training inputs. Validation was modality-specific, with separate MRI and CT splits, while testing involved 60 cases of unpaired CT and MRI images from AMOS2020. 3D Dice scores were computed after reconstructing the predicted slices into volumes. Table \ref{tab:dataset_stats_reformatted} summarizes the number of 3D volumes and 2D slices used per organ.

\begin{table}[h!]
\centering
\caption{Number of cases in the training and validation sets for each organ in the TotalSegmentator dataset. The numbers in parentheses indicate the total number of slices available for each organ in the respective modality.}
\label{tab:dataset_stats_reformatted}
\begin{tabular}{l c @{\hspace{1.5em}} c @{\hspace{2em}} c @{\hspace{1.5em}} c}
\hline
 & \multicolumn{2}{c}{\textbf{Training}} & \multicolumn{2}{c}{\textbf{Validation}} \\
 \textbf{Organ} & \textbf{MRI} & \textbf{CT} & \textbf{MRI} & \textbf{CT} \\ 
 \hline
Liver          & 80 (28,142)   & 704 (112,569) & 20 (2,754)    & 177 (25,744) \\
Kidneys        & 84 (28,142)   & 662 (112,018) & 22 (2,865)    & 166 (24,648) \\
Gall Bladder   & 65 (22,701)   & 510 (90,805)  & 17 (2,330)    & 128 (20,556) \\
Spleen         & 81 (27,796)   & 687 (111,183) & 21 (2,899)    & 172 (25,189) \\
Pancreas       & 76 (26,985)   & 631 (107,938) & 19 (2,611)    & 158 (23,022) \\
\hline
\end{tabular}
\end{table}

\subsection{Implementation and Experimental setup}

We implemented our proposed FedGIN using \texttt{PyTorch}, with FL handled by the \texttt{Flower} framework \cite{beutel2020flower}. All experiments were run on an \texttt{NVIDIA A40} GPU (48 GB VRAM) using \texttt{CUDA 12.4}. 
We used a 2D U-Net \cite{ronneberger2015unetconvolutionalnetworksbiomedical} as the base segmentation model, with modifications such as strided convolutions for downsampling and bilinear interpolation for upsampling, implemented explicitly via \texttt{F.interpolate} within the decoder path. This ensures accurate spatial alignment between encoder and decoder feature maps during skip connections and maintains resolution consistency in the final output. LeakyReLU activations, batch normalization, and dropout (\(p=0.3\), \(p=0.4\) in the bottleneck) were used throughout. The final layer applies a sigmoid activation function to produce binary segmentation masks. All convolutional layers in the network are initialized using Kaiming He initialization \cite{he2015delvingdeeprectifierssurpassing}. Training used a combined Focal loss and Dice loss, weighted equally, and was optimized using \texttt{AdamW} (learning rate \(5 \times 10^{-4}\), weight decay \(1 \times 10^{-4}\)). To improve training efficiency, we used a learning rate scheduler that reduces the learning rate when the validation performance plateaus, along with automatic mixed precision to accelerate computation and reduce memory usage. In federated training, each client (CT and MRI) trained locally for 1 epochs per round across 100 communication rounds using FedAvg \cite{mcmahan2017communication}. GIN augmentation was applied on-the-fly at each client. For centralized training, we trained the same U-Net model for 100 epochs using the combined MRI and CT data. 3D Dice scores were computed by stacking 2D predictions into full patient volumes.

\subsection{Results and Analysis}

\begin{figure}[t]
\centering
\includegraphics[width=0.9\textwidth]{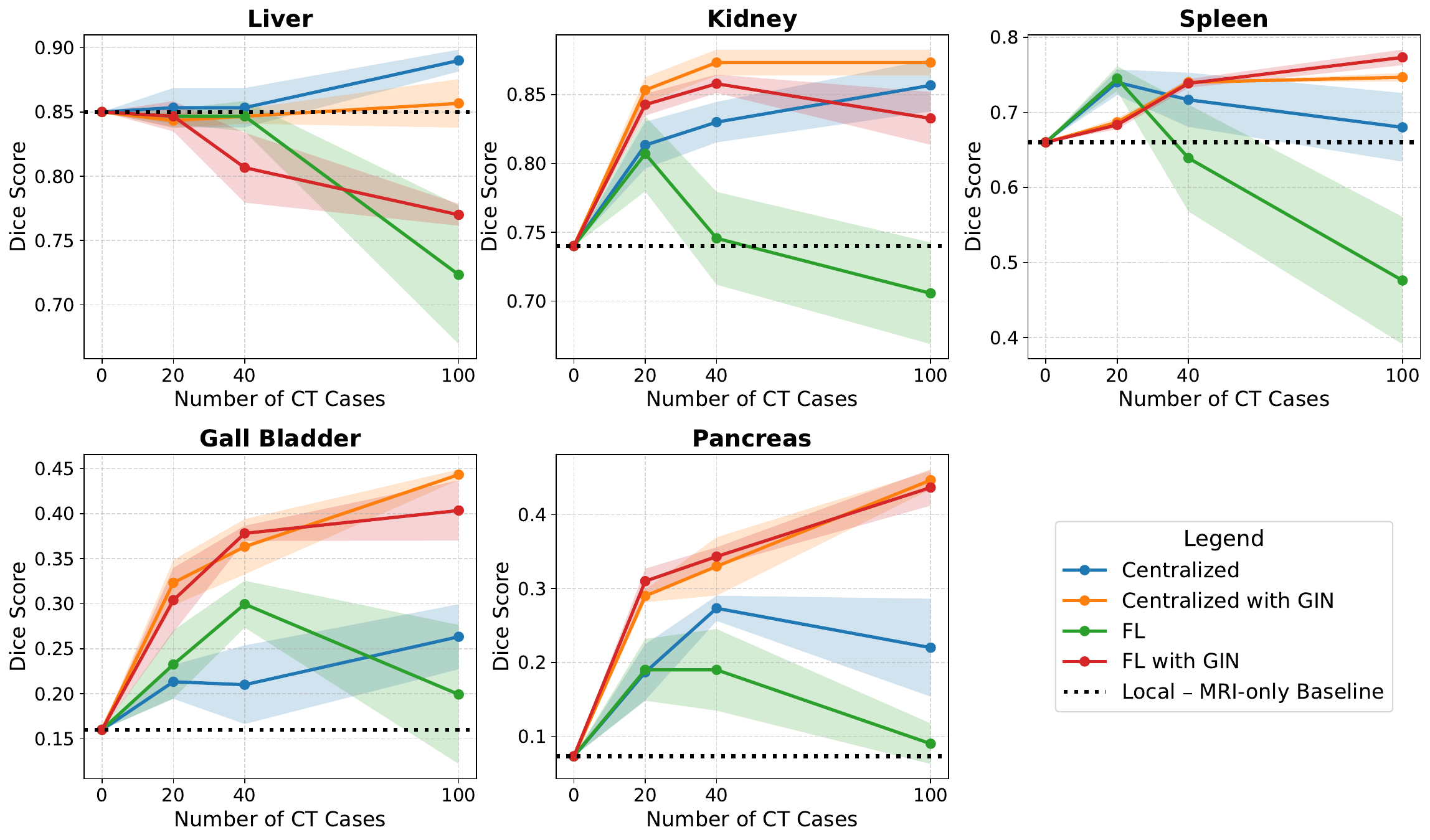}
\caption{3D Dice scores on AMOS MRI test set across increasing CT cases. Comparison of local MRI-only training, centralized models with/without GIN, and FL with/without GIN (FedGIN).}
\label{fig2}

\end{figure}

We began by evaluating our FedGIN approach in a single-client setting under limited data conditions. For each organ, we first trained a baseline model using 20 MRI cases and then progressively added 20, 40, and 100 CT cases to examine whether incorporating a complementary modality can improve segmentation performance. We compared four training configurations: centralized training with and without GIN, FL with MRI and CT clients without GIN, and our proposed FedGIN setup with GIN. All experiments used the same hyperparameters and were repeated with three random seeds for fair comparison. Figure~\ref{fig2} displays the 3D Dice scores on the AMOS MRI test set for each organ, with the dashed line representing the local MRI-only baseline. Adding CT data improved MRI segmentation performance for most organs, except the liver. The improvement was most notable in low-contrast, complex organs like the spleen, gallbladder, and pancreas, which are challenging for segmentation tasks. In these cases, FedGIN and centralized training with GIN performed similarly, surpassing other configurations. For the spleen and pancreas, FedGIN closely matched centralized GIN models, demonstrating effective domain generalization in decentralized settings. In the case of the liver and kidneys, where anatomical consistency and intensity distribution are more homogeneous, centralized GIN training showed a slight performance advantage over FedGIN. Notably, for the liver, FedGIN did not improve performance over the local baseline or centralized counterparts. This underperformance may be due to the saturation of domain-specific characteristics in liver segmentation, where MRI-only models are already near their maximum performance. Additionally, the relatively small benefit of CT-derived variation for such organs may limit the marginal gain from GIN augmentation in the federated setting. In contrast, FL without GIN consistently underperformed and further degraded as more CT data was added. This highlights the negative effect of unharmonized modality mixing and confirms that GIN stabilizes the performance in this cross-domain federated setup. These findings support our hypothesis under specific conditions: when segmenting structurally complex and low-contrast organs (e.g., spleen, gall bladder, pancreas), the incorporation of CT data significantly enhances MRI segmentation performance. Moreover, under heterogeneous data distributions across clients, GIN augmentation proves essential for closing the performance gap between centralized and federated training. In contrast, for organs that are simpler to segment, such as the liver and kidneys, where MRI data alone already yields high baseline performance, the added benefit of CT data and GIN is more limited.

\begin{figure} [t]
\centering
\includegraphics[width=0.9\textwidth]{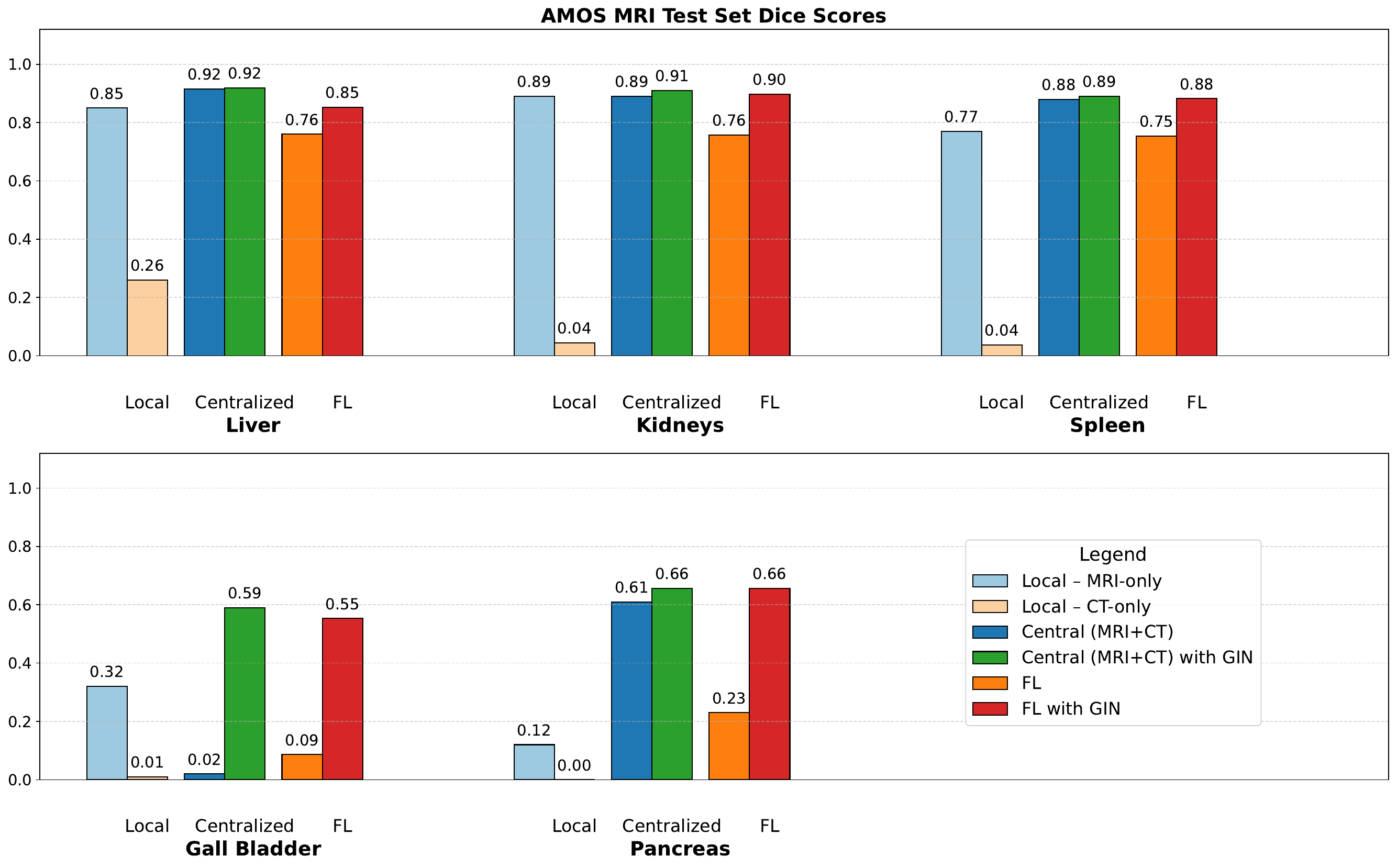}
\caption{Comparison of Dice Scores from AMOS MRI Test Set across TotalSegmentator-Trained Local, Centralized, and Federated Learning Models, with and without GIN, for different organs.} \label{fig3}

\end{figure}

\begin{figure}[t]
\centering
\includegraphics[width=0.9\textwidth]{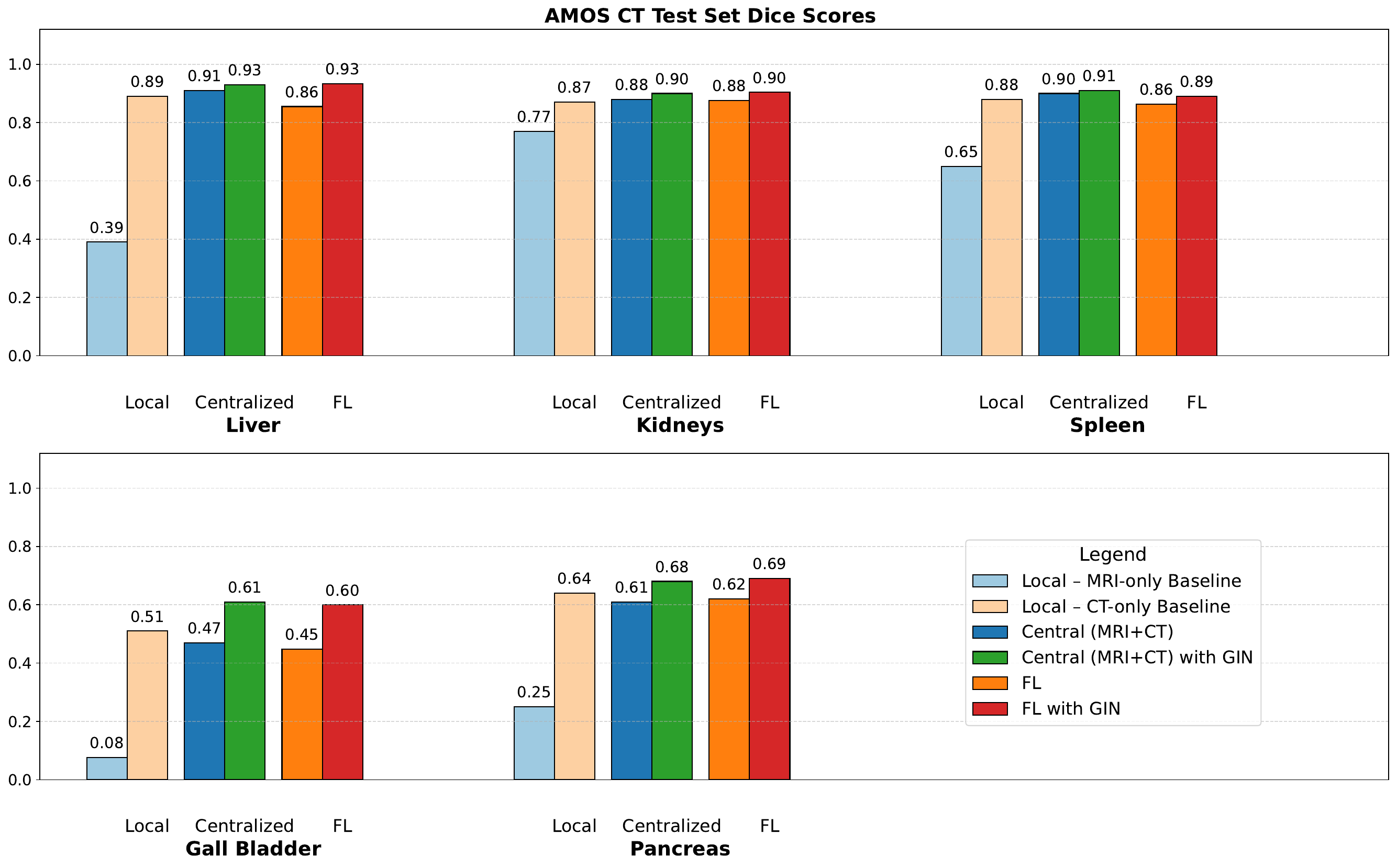}
\caption{Comparison of Dice Scores from AMOS CT Test Set across TotalSegmentator-Trained Local, Centralized, and Federated Learning Models, with and without GIN, for different organs.} \label{fig4}

\end{figure}

To assess the generalizability of our method at scale, we trained models using the complete TotalSegmentator dataset for each organ, as detailed in Table~\ref{tab:dataset_stats_reformatted}. These models were then evaluated on the AMOS CT and MRI test sets. We compared five training configurations: local models trained on MRI and CT data separately, centralized training on combined CT and MRI data (with and without GIN), and FL with separate MRI and CT clients (with and without GIN). Figures~\ref{fig3} and~\ref{fig4} show that models trained on unpaired multimodal data outperformed single-modality baselines in both test sets. GIN-centralized and FedGIN models achieved the best performance across all organs. In the AMOS CT test set, FedGIN and centralized GIN had comparable Dice scores in all five organs, with pancreas segmentation reaching 0.69 for FedGIN and 0.68 for centralized GIN. For gallbladder segmentation, FedGIN improved the score from 0.08 (MRI-only) and 0.51 (CT-only) to over 0.60 with multimodal training. This highlights the effectiveness of integrating CT and MRI during training, even in a federated setting. Similarly, on the AMOS MRI test set, both FedGIN and centralized GIN achieved strong and balanced generalization. In challenging cases like the gallbladder and pancreas, performance increased significantly compared to local models trained only on CT or MRI. These improvements confirm that GIN-harmonized cross-modal training leads to better generalization in both centralized and federated settings. GIN multimodal training not only outperformed local baselines but also demonstrated that FedGIN can match the performance of centralized training without requiring centralized data access.

These results show that multimodal training consistently outperformed the local baseline models across all organs. Both centralized and federated models with GIN achieved higher Dice scores, demonstrating effective modality alignment and improved generalization. Notably, FL with GIN closely matched centralized GIN performance, validating its robustness under decentralized training. Recent studies \cite{d2024totalsegmentator,ciausu2024towards} report the Dice scores of approximately 0.83 for the liver and up to 0.91 for the kidneys, which closely align with our results. Notably, our approach achieves comparable performance without relying on complex network architectures or centralized access to multimodal data. Instead, we employ a lightweight GIN augmentation strategy within an FL framework, offering a scalable and privacy-preserving alternative. Furthermore, our method addresses persistent segmentation challenges in low-contrast and structurally variable organs, such as the pancreas, spleen, and gallbladder, more effectively than local baseline models. These results show our framework as demonstrating competitive performance under realistic deployment constraints. As such, our work contributes a strong and generalizable baseline for future multimodal segmentation research in federated settings.

\section{Conclusion}

In this work, we introduced a federated multimodal organ segmentation framework that integrates unpaired CT and MRI data from multiple clients, enhanced with GIN augmentation to mitigate domain shifts. Our experiments, conducted on both small-scale and large-scale datasets, consistently demonstrate that incorporating CT data significantly enhances segmentation performance on MRI, particularly for organs that present challenging segmentation tasks, such as the pancreas and gallbladder. FedGIN shows competitive results compared to centralized training, demonstrating its potential for cross-modality learning in a federated setting. Furthermore, our large-scale evaluations reinforce the advantages of multimodal training with GIN, highlighting its ability to generalize across domains. These findings support the feasibility of scalable, federated, and modality-agnostic learning strategies for real-world clinical segmentation tasks.

\bibliographystyle{splncs04} %
\bibliography{ref}

\begin{thebibliography}{10}
\providecommand{\url}[1]{\texttt{#1}}
\providecommand{\urlprefix}{URL }
\providecommand{\doi}[1]{https://doi.org/#1}

\bibitem{beutel2020flower}
Beutel, D.J., Topal, T., Mathur, A., Qiu, X., Fernandez-Marques, J., Gao, Y., Sani, L., Li, K.H., Parcollet, T., de~Gusm{\~a}o, P.P.B., et~al.: Flower: A friendly federated learning research framework. arXiv preprint arXiv:2007.14390  (2020)

\bibitem{chen2025generalizable}
Chen, B., Zhu, Y., Ao, Y., Caprara, S., Sutter, R., R{\"a}tsch, G., Konukoglu, E., Susmelj, A.: Generalizable single-source cross-modality medical image segmentation via invariant causal mechanisms. In: 2025 IEEE/CVF Winter Conference on Applications of Computer Vision (WACV). pp. 3592--3602. IEEE (2025)

\bibitem{ciausu2024towards}
Ciausu, C., Krishnaswamy, D., Billot, B., Pieper, S., Kikinis, R., Fedorov, A.: Towards automatic abdominal mri organ segmentation: Leveraging synthesized data generated from ct labels. arXiv preprint arXiv:2403.15609  (2024)

\bibitem{d2024totalsegmentator}
D'Antonoli, T.A., Berger, L.K., Indrakanti, A.K., Vishwanathan, N., Wei{\ss}, J., Jung, M., Berkarda, Z., Rau, A., Reisert, M., K{\"u}stner, T., et~al.: Totalsegmentator mri: Sequence-independent segmentation of 59 anatomical structures in mr images. arXiv preprint arXiv:2405.19492  (2024)

\bibitem{guan2024federated}
Guan, H., Yap, P.T., Bozoki, A., Liu, M.: Federated learning for medical image analysis: A survey. Pattern Recognition p. 110424 (2024)

\bibitem{he2015delvingdeeprectifierssurpassing}
He, K., Zhang, X., Ren, S., Sun, J.: Delving deep into rectifiers: Surpassing human-level performance on imagenet classification (2015), \url{https://arxiv.org/abs/1502.01852}

\bibitem{ji2022amos}
Ji, Y., Bai, H., Ge, C., Yang, J., Zhu, Y., Zhang, R., Li, Z., Zhanng, L., Ma, W., Wan, X., et~al.: Amos: A large-scale abdominal multi-organ benchmark for versatile medical image segmentation. Advances in neural information processing systems  \textbf{35},  36722--36732 (2022)

\bibitem{lassau2020three}
Lassau, N., Bousaid, I., Chouzenoux, E., Lamarque, J.P., Charmettant, B., Azoulay, M., Cotton, F., Khalil, A., Lucidarme, O., Pigneur, F., et~al.: Three artificial intelligence data challenges based on ct and mri. Diagnostic and Interventional Imaging  \textbf{101}(12),  783--788 (2020)

\bibitem{mcmahan2017communication}
McMahan, B., Moore, E., Ramage, D., Hampson, S., y~Arcas, B.A.: Communication-efficient learning of deep networks from decentralized data. In: Artificial intelligence and statistics. pp. 1273--1282. PMLR (2017)

\bibitem{myrzashova2025bcftl}
Myrzashova, R., Alsamhi, S.H., Shvetsov, A.V., Hawbani, A., Guizani, M., Wei, X.: Bcftl: Blockchain-enabled multimodal federated transfer learning for decentralized alzheimer’s diagnosis. IEEE Internet of Things Journal  (2025)

\bibitem{ouyang2022causality}
Ouyang, C., Chen, C., Li, S., Li, Z., Qin, C., Bai, W., Rueckert, D.: Causality-inspired single-source domain generalization for medical image segmentation. IEEE Transactions on Medical Imaging  \textbf{42}(4),  1095--1106 (2022)

\bibitem{pati2024privacy}
Pati, S., Kumar, S., Varma, A., Edwards, B., Lu, C., Qu, L., Wang, J.J., Lakshminarayanan, A., Wang, S.h., Sheller, M.J., et~al.: Privacy preservation for federated learning in health care. Patterns  \textbf{5}(7) (2024)

\bibitem{raggio2025fedsynthct}
Raggio, C.B., Zabaleta, M.K., Skupien, N., Blanck, O., Cicone, F., Cascini, G.L., Zaffino, P., Migliorelli, L., Spadea, M.F.: Fedsynthct-brain: A federated learning framework for multi-institutional brain mri-to-ct synthesis. Computers in Biology and Medicine  \textbf{192},  110160 (2025)

\bibitem{ronneberger2015unetconvolutionalnetworksbiomedical}
Ronneberger, O., Fischer, P., Brox, T.: U-net: Convolutional networks for biomedical image segmentation (2015), \url{https://arxiv.org/abs/1505.04597}

\bibitem{wang2025multi}
Wang, N., Deng, Y., Fan, S., Yin, J., Ng, S.K.: Multi-modal one-shot federated ensemble learning for medical data with vision large language model. arXiv preprint arXiv:2501.03292  (2025)

\bibitem{wang2025clusmfl}
Wang, X., Zhou, R., Xie, H., Tang, X., He, L., Yang, C.: Clusmfl: A cluster-enhanced framework for modality-incomplete multimodal federated learning in brain imaging analysis. arXiv preprint arXiv:2502.12180  (2025)

\bibitem{wasserthal2023totalsegmentator}
Wasserthal, J., Breit, H.C., Meyer, M.T., Pradella, M., Hinck, D., Sauter, A.W., Heye, T., Boll, D.T., Cyriac, J., Yang, S., et~al.: Totalsegmentator: robust segmentation of 104 anatomic structures in ct images. Radiology: Artificial Intelligence  \textbf{5}(5),  e230024 (2023)

\bibitem{yuan2024communication}
Yuan, L., Han, D.J., Wang, S., Upadhyay, D., Brinton, C.G.: Communication-efficient multimodal federated learning: Joint modality and client selection. arXiv preprint arXiv:2401.16685  (2024)

\bibitem{zhao2018federated}
Zhao, Y., Li, M., Lai, L., Suda, N., Civin, D., Chandra, V.: Federated learning with non-iid data. arXiv preprint arXiv:1806.00582  (2018)

\end{thebibliography}

\end{document}